\documentclass{article}

\usepackage{microtype}
\usepackage{graphicx}
\usepackage{subcaption}
\usepackage{booktabs} % for professional tables

\usepackage{hyperref}

\usepackage[accepted]{icml2021}

\usepackage{array}
\newcolumntype{P}[1]{>{\centering\arraybackslash}p{#1}}

\usepackage{booktabs}

\usepackage{amsmath,amsfonts,amssymb,amsthm}
\usepackage{mathtools}
\usepackage{commath}

\theoremstyle{definition}

\DeclarePairedDelimiterX{\infdivx}[2]{(}{)}{%
  #1\;\delimsize\|\;#2%
}

\usepackage{float}
\usepackage{adjustbox}
\usepackage{wrapfig}
\usepackage{multirow}
\definecolor{lightgrey}{rgb}{0.43,0.43,0.43}

\definecolor{deepmagenta}{rgb}{0.8, 0.0, 0.8}

\icmltitlerunning{The Impact of Negative Sampling on Contrastive Structured World Models}

\begin{document}

\twocolumn[
\icmltitle{The Impact of Negative Sampling on Contrastive Structured World Models}




\begin{icmlauthorlist}
\icmlauthor{Ondrej Biza}{nu}
\icmlauthor{Elise van der Pol}{am}
\icmlauthor{Thomas Kipf}{brain}
\end{icmlauthorlist}

\icmlaffiliation{nu}{Northeastern University, Boston, MA, USA}
\icmlaffiliation{am}{University of Amsterdam, Netherlands}
\icmlaffiliation{brain}{Google Research, Brain Team}

\icmlcorrespondingauthor{Ondrej Biza}{biza.o@northeastern.edu}

\icmlkeywords{Contrastive Learning, World Models, Arcade Learning Environment, Self-Supervised Learning}

\vskip 0.3in
]



\printAffiliationsAndNotice{}  % leave blank if no need to mention equal contribution

\begin{abstract}
World models trained by contrastive learning are a compelling alternative to autoencoder-based world models, which learn by reconstructing pixel states. In this paper, we describe three cases where small changes in how we sample negative states in the contrastive loss lead to drastic changes in model performance. In previously studied Atari datasets, we show that leveraging time step correlations can double the performance of the Contrastive Structured World Model. We also collect a full version of the datasets to study contrastive learning under a more diverse set of experiences.
\end{abstract}

\section{Introduction}
\label{sec:intro}
Representing dynamic environments in terms of objects and their relations is a powerful approach to dealing with unseen configurations and hereby provides a path towards improving generalization and data efficiency. Self-supervised methods for learning object-centric representations of a visual scene have to date primarily employed reconstruction-based objectives~\cite{greff2019multi,engelcke2019genesis,burgess2019monet,janner19,veerapaneni19,kossen20}. More recent work has explored the use of contrastive learning in this context~\cite{kipf20,racah2020slot,lowe2020learning}. Contrastive approaches learn representations entirely in latent space by moving encodings of similar entities (positive samples) close together, and moving encodings of dissimilar entities (negative samples) away from each other. The advantage of contrastive coding is that there is no longer a need to reconstruct the entire visual scene. As a result, we do not have to encode unnecessary visual details in the latent code, nor do we have to train a decoder network. 

When training contrastive coding methods, we generally know how to select positive samples, but it is less obvious how to select negative samples. When learning representations of images, data-augmented versions of the same image are commonly used as positive samples~\cite{dosovitskiy2015discriminative, chen2020simple, wu2018unsupervised, federici2020learning}. In world models, states that transition into each other should also be close together in latent space~\cite{oord2018representation, kipf20, vdp20}. The challenge in selecting negative samples is that there is a trade-off between samples that are too dissimilar and samples that are too similar. 
%When For example, sampling a different image at random can lead to using the same class as a negative sample.  
In world models, sampling negative states at random from a replay buffer may make it trivial to differentiate between positive and negative examples. For example, by only encoding the set of objects in a scene without encoding their positions. By contrast, when using negative samples from the same episode, the learned model may not encode time-invariant features that are specific to an episode, which may give rise to different long term rewards. 
%\elise{these examples may only be clear after reading the experiment section} O: I see what you mean, but I can't think of better ones right now.
%since these features will be identical in positive and negative examples.
%Similarly, in world models sampling different states at random from the replay memory or a different episode can lead to accidentally using the same abstract state as a negative sample. 
%In practice, we therefore need to select negative examples that allow a model to distinguish between states that are similar but lead to different long term rewards.

In this paper, we demonstrate that the negative sampling heuristic can drastically influence performance for object-centric contrastive learning approaches. To do so, we apply a variety of sampling heuristics to train contrastive structured world models (C-SWMs, \citet{kipf20}), a recently proposed contrastive method for learning object-based representations.  
Our contributions are:

\textbf{\#1:} We show that leveraging the correlation of states in a particular time step of an episode when sampling negative examples can double the prediction accuracy relative to \citet{kipf20} in Atari games.

\textbf{\#2:} We identify (and propose a solution to) a problem where correlation of within-episode states leads to a contrastive model failing to represent objects.

\textbf{\#3:} We extend the Atari datasets used in \citet{kipf20} to cover a diverse set of states within the games. We discuss evaluation problems on these datasets.

\section{Contrastive Structured World Model}
\label{sec:background}

C-SWMs model transition dynamics of fully-observable image-based environments. A C-SWM employs a representation that factorizes into slots for individual objects. A state $s_t$ at time step $t$, represented as an image, is first encoded into a latent matrix with $K$ slots $z_t^{1:K}$. Then, the C-SWM uses an action-conditioned Graph Neural Network (GNN) to model interaction between object slots. 

The transition GNN receives a sequence of $K$ object encodings $z_t^{1:K}$, where $z_t^i \in \mathbb{R}^D$. Optionally, it receives an action $a_t$. The network models interactions between objects through an edge neural network $f_\text{edge}$:
\begin{align}
    e_t^{(i,j)} = f_\text{edge}(z_t^i, z_t^j; \theta).
\end{align}
The edge encodings $e_t^{(i,j)}$ are then used to predict the next state of each object. The node network $f_\text{node}$ sums over the edge encodings, ignoring self-loops, and predicts the difference between the current and next state of each object
\begin{align}
    \hat{z}_{t+1}^i = z_{t}^i + f_\text{node}(z_{t}^i, a_t, \sum_{i \neq j} e_t^{(i,j)}).
\end{align}
Unlike comparable factored world models \cite{janner19,veerapaneni19,kossen20}, a C-SWM does not rely on rendering of images to reconstruct inputs. Instead, a C-SWM is trained end-to-end using a contrastive loss, which is defined in the latent space. This loss minimizes the distance between the predicted next states $\hat{z}_{t+1}$ and a positive sample $z_{t+1}$ whilst ensuring that the states $z_t$ are distinct from negative samples $\bar{z}_t$,
\begin{align}
    \mathcal{L} = H(\hat{z}_{t+1}, z_{t+1}) + \max(0, \gamma - H(\bar{z}_t, z_t)).
\end{align}
In this loss, the function $H$ is the average distance between the object encodings in a pair of states,
\begin{align}
    H(z, z') = \frac{1}{2 K \sigma^2} \sum_{k=1}^K || z^k - z'^k ||^2_2.
\end{align}
To minimize the loss, the model's prediction of the next state $\hat{z}_{t+1}$ has to be near the actual next state $z_{t+1}$. At the same time, the randomly sampled negative state $\bar{z}_t$ has to be separated from the current states $z_t$ by at least $\gamma$.

\section{Negative Sampling Strategies}
\label{sec:methods}

We compare three strategies for selecting negative samples:

\textbf{Baseline Negatives (C-SWM).} The baseline sampling strategy in the original C-SWM is to select 1024 positive examples by sampling transitions using a replay buffer and to select negative examples randomly from the set of 1023 non-paired states at random, resulting in negative samples that are mostly uncorrelated with the positive sample.

%\elise{we may need to add a short motivation for why we consider time-alignment}

The baseline sampling strategy often samples negative states that are too easy to distinguish. For example, two states in the game of Pong where the player paddle is on the opposite sides of the screen do not require the model to have a precise understanding of the position of the paddles. 

\textbf{Time-Aligned Negatives (C-SWM-TA).} For each state from episode $e$ at time step $t$, we sample a negative state from a randomly chosen episode $e_r$, $e_r \neq e$, at time step $t$. I.e. the negative states come from different episodes but same time steps as the positive states.

The above heuristic is only meaningful when time step carries information. That is the case in Section \ref{sec:exp:atari_orig}, where time step 0 is at the start of a game. But, it fails if the episode can start at any point in the game (Section \ref{sec:exp:atari_full}) or if all actions are reversible (Section \ref{sec:exp:shapes}). Hence, we explore a third strategy for sampling challenging negative states.

\textbf{Episodic and Out-of-episode Negatives (C-SWM-ER).} To select negative samples that will be more difficult to distinguish from positive examples, we construct a mixture of within-episode and out-of-episode negatives. With probability $\beta$, we sample a negative state for episode $e$ at time $t$ by selecting a random time $t_r \neq t$ within the same episode. With probability $1-\beta$, we select a negative sample from a different episode at a random time step. 

%Two instantiations of the above strategy are $\beta=0$ and $\beta=0.5$. With $\beta=0$, we only sample negative states from different episodes, but the same time step as the positive sample. With $\beta=0.5$, we have an equal mix of negative states from the same episode and from different episodes.

% V1
%\textbf{Mixed Episodic and Random Negatives (ER)}:  %\tk{I think we should try to find a more descriptive name other than "new negatives". If the name would capture what the actual difference is (e.g. across episode sampling, within episode sampling vs. mixed sampling), that'd be great.}: 
%To construct negative samples that cannot be trivially distinguished from positive examples, we sample negative and positive states from a single episode. With probability $\gamma$, we sample a negative state for episode $e$ at time $t$ by selecting a random time $t_r \neq t$ within the same episode. With probability $1-\gamma$, sample a negative state at random from a different episode, which results in negative samples similar to those in the baseline strategy.
%\textbf{Mixed Episodic and Time-Aligned Negatives (EA).} In addition to mixing episodic negatives with baseline random negatives, we consider a sampling strategy in which we select an episodic negative with probability $\gamma$ as above, and with probability $1-\gamma$ we select a negative sample from the same time $t$ in a different episode.

\section{Experiments}
\label{sec:exp}

We present three case studies where changes in negative sampling have a surprisingly large impact. The source code is located at \url{https://github.com/ondrejba/negative-sampling-icml-21}.

\begin{figure}[h]
    \centering
    \begin{subfigure}{0.24\linewidth}
        \centering
        \includegraphics[width=1.0\linewidth]{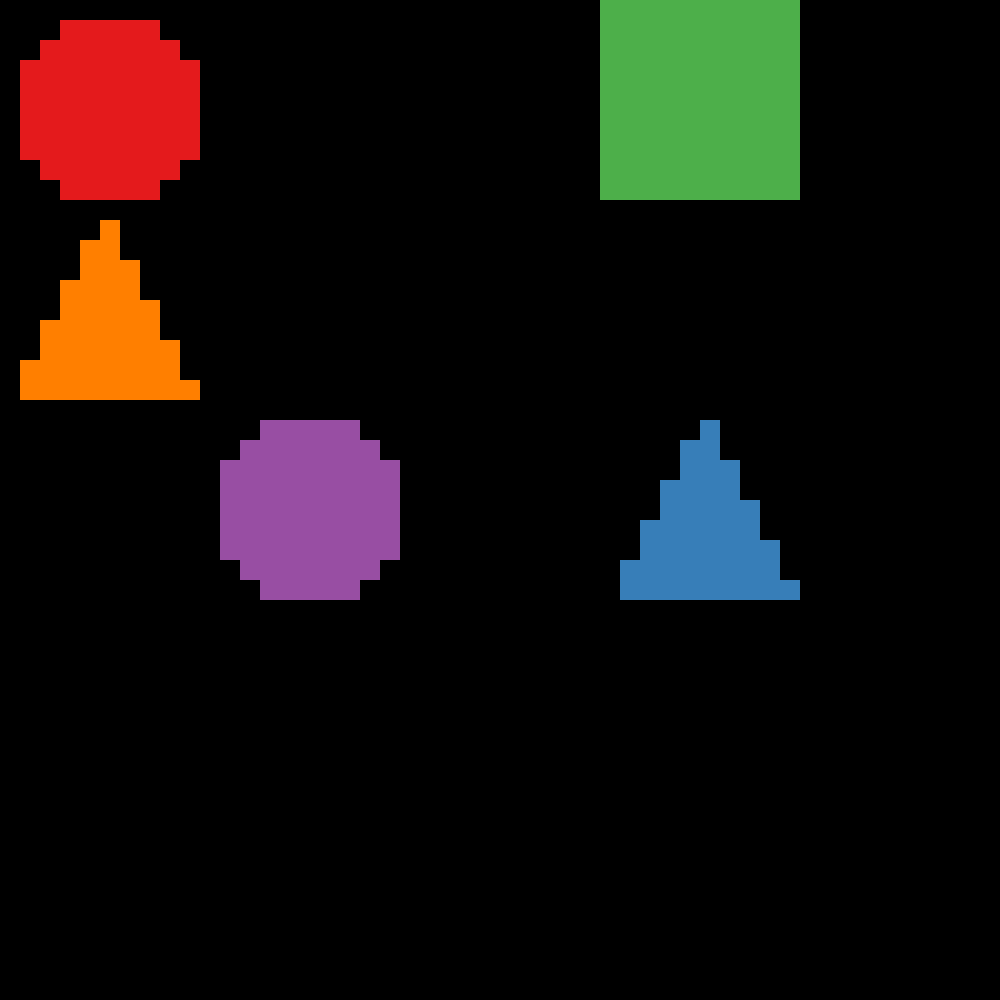}
    \end{subfigure}
    \begin{subfigure}{0.24\linewidth}
        \centering
        \includegraphics[width=1.0\linewidth]{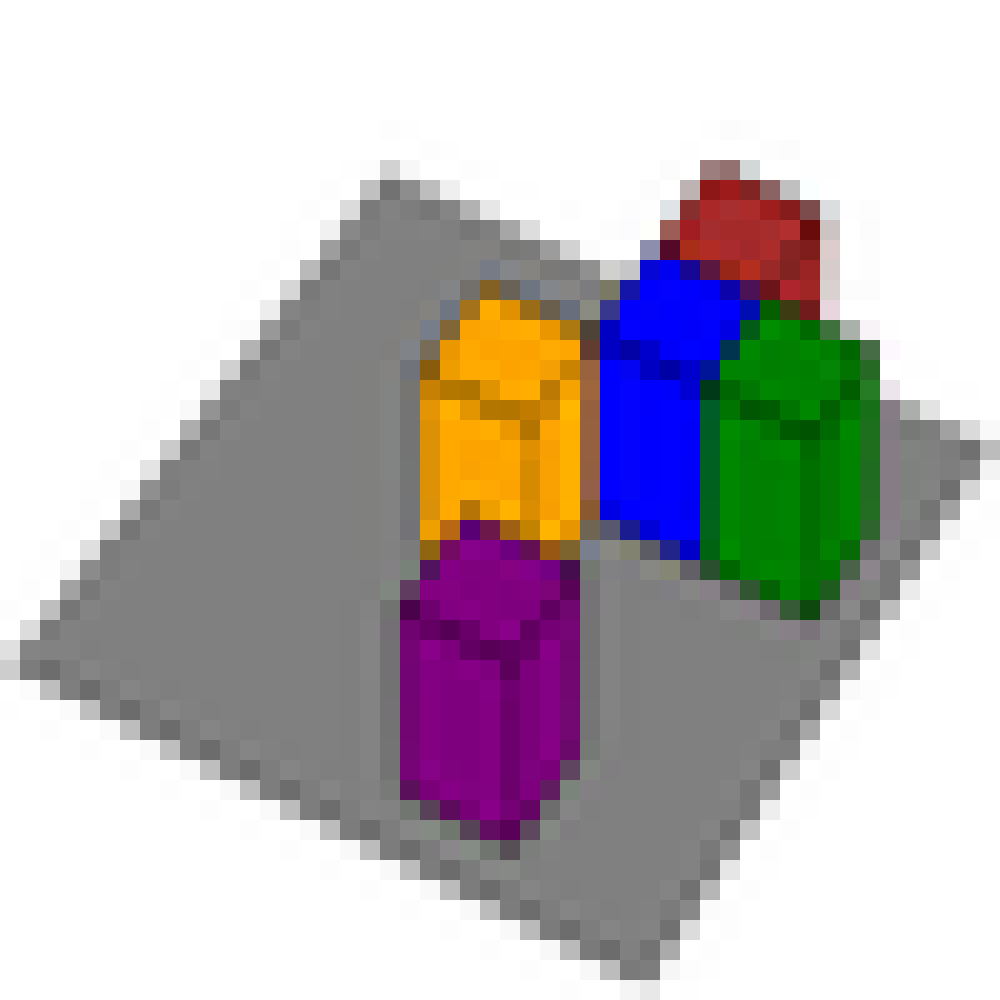}
    \end{subfigure}
    \begin{subfigure}{0.24\linewidth}
        \centering
        \includegraphics[width=1.0\linewidth]{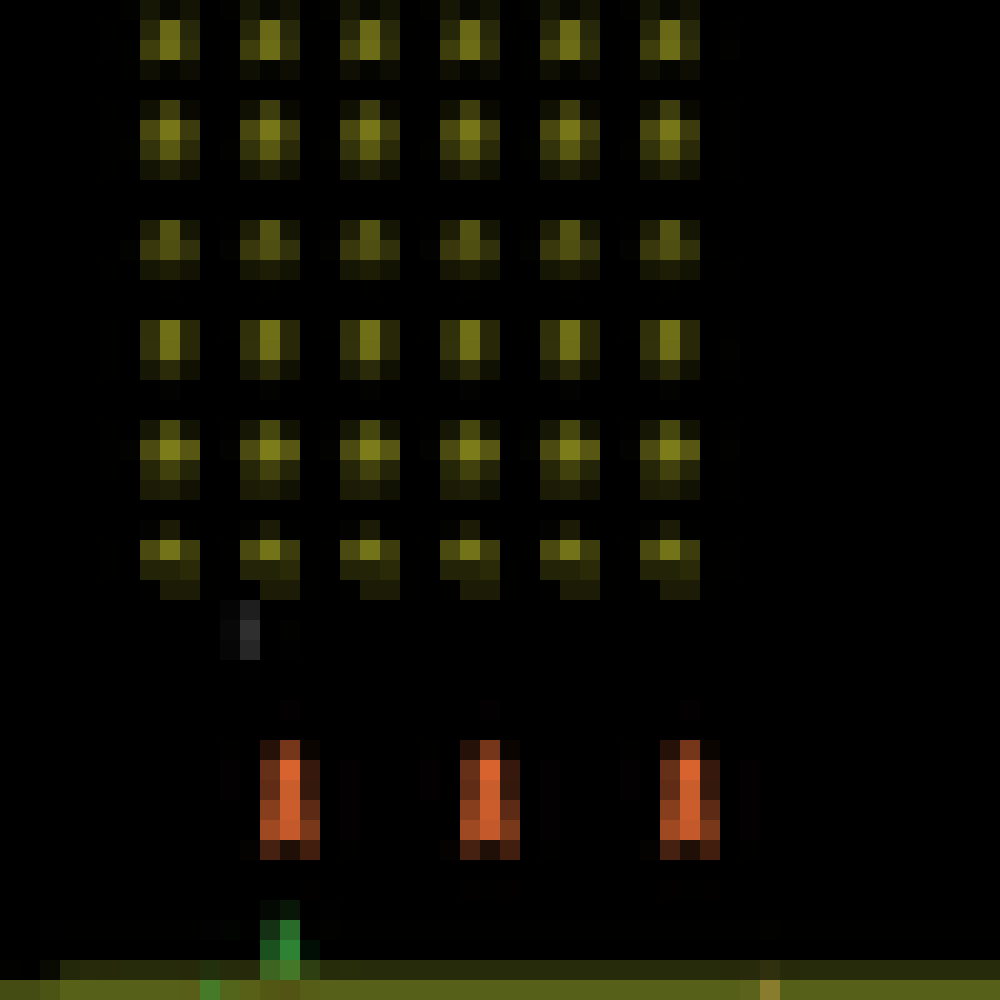}
    \end{subfigure}
    \begin{subfigure}{0.24\linewidth}
        \centering
        \includegraphics[width=1.0\linewidth]{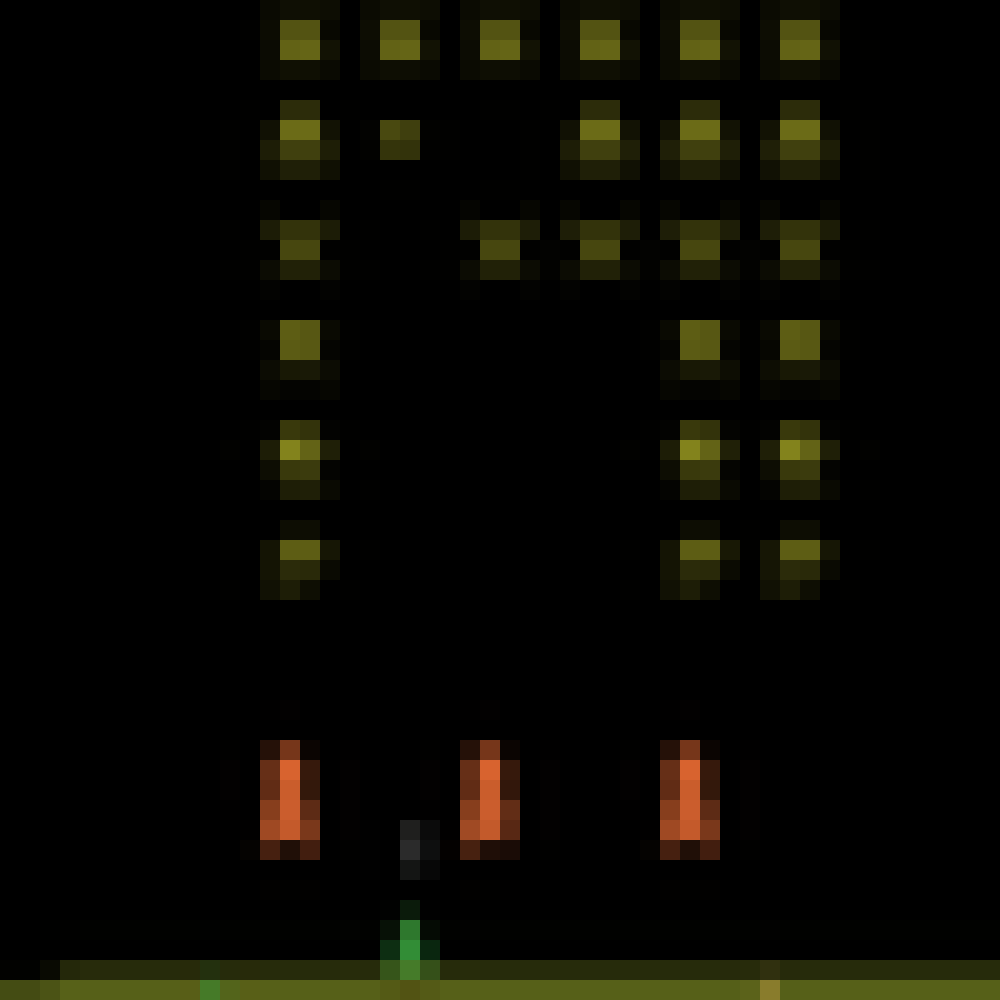}
    \end{subfigure}
    \caption{Examples states of environments we use. From left to right: 2D Shapes, 3D Cubes and Space Invaders.}
    \label{fig:envs}
\end{figure}

% \subsection{Datasets}
% \label{sec:exp:data}

% \textbf{2D/3D Grid Immovable} are the Shapes and Cube grid-world environments from \cite{kipf20} with our modifications. There are five distinct objects in a $5{\times}5$ grid-world that can be moved in the four cardinal directions. We demonstrate the problem with baseline negative sampling by making the first two objects (red and blue objects) immovable. They are initialized randomly at the start of each (training and validation) episode, but cannot be moved.

% \textbf{Atari Pong and Space Invaders} are datasets collected in the games Pong and Space Invaders using the Arcade Learning Environment \cite{bellemare2013arcade}. Following \citet{kipf20}, we skip the first 50 - 60 frames of the games (the game does not start instantly) and then collect a sequence of ten transitions with random actions.

% \textbf{Full Atari Pong and Space Invaders} are new datasets we collected to evaluate C-SWM on a more diverse set of experiences. We use a trained model-free RL agent with an $\epsilon$-greedy exploration policy to play each game. After a randomized number of time steps, the agent stops and we collect a sequence of ten transitions with random actions. The episodes in our dataset start at different points in the games (e.g. with a different number of aliens shot in Space Invaders), but we do not introduce any bias in action selection by using random actions.

\subsection{Episodic Correlations in 2D/3D Grid Immovable}
\label{sec:exp:shapes}

\begin{figure}[h]
    \centering
    \begin{subfigure}{0.49\linewidth}
        \centering
        \includegraphics[width=1.0\linewidth]{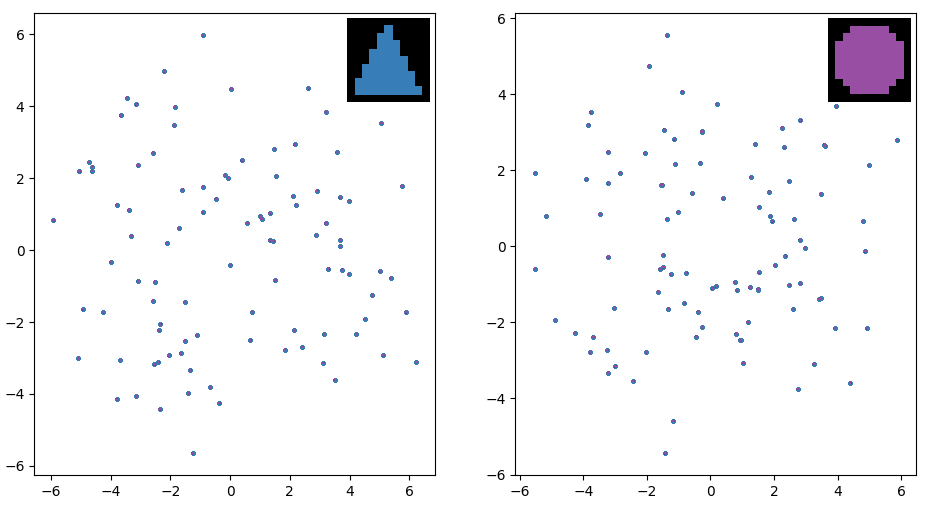}
        \caption{}
    \end{subfigure}
    \begin{subfigure}{0.49\linewidth}
        \centering
        \includegraphics[width=1.0\linewidth]{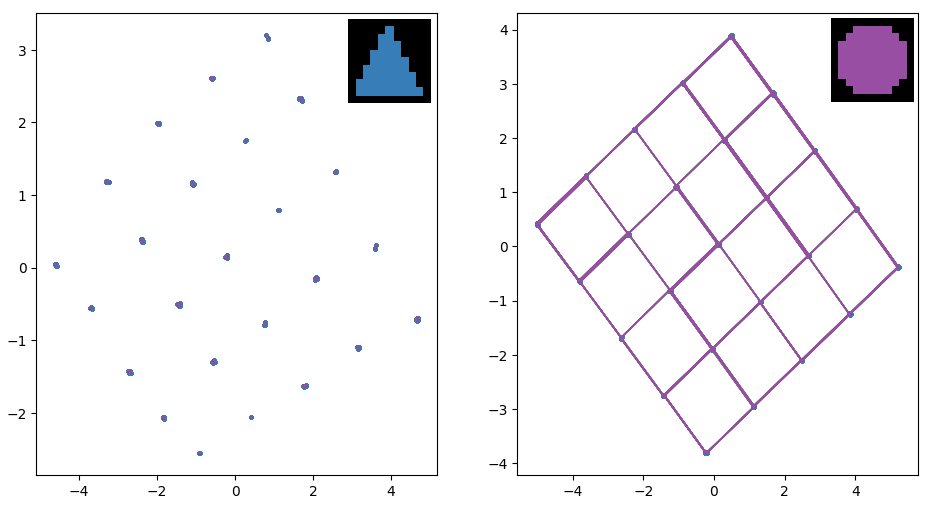}
        \caption{}
    \end{subfigure}
    \vspace{-0.5\baselineskip}
    \caption{Qualitative difference between the learned latent space of (a) baseline C-SWM and (b) C-SWM-ER ($\beta=0.5$) in the 2D grid-world. Points are 2D encodings of the blue triangle (immovable) and purple circle (movable) from 5k states sampled from the evaluation set. Lines are predicted next states given random actions. C-SWM-ER learns both the state of the immovable blue triangle and the movement of the purple circle.}
    \label{fig:shapes_latent}
    \vspace{-0.5\baselineskip}
\end{figure}

\textbf{Setup.} We start by demonstrating that a simple change to the two grid-world environments (2D Shapes and 3D Cubes; Figure \ref{fig:envs}) used in \citet{kipf20} can cause C-SWM to fail completely. In both 2D Shapes and 3D Cubes, five objects are randomly initialized in a $5{\times}5$ grid-world at the start of each episode. In the original environments, all five objects can be moved in the four cardinal directions. We simply make the first two objects (red and blue) immovable.

\textbf{Result.} Immovable objects causes the Hits @1 score for predictions 10 steps forward to drop from 100\% to 12\% for the baseline C-SWM (Table \ref{tab:ap:all_1}, first row). Qualitatively, we notice that the model does not model transitions at all. Instead, it only focuses on detecting the positions of the two immovable objects (Figure \ref{fig:shapes_latent_slots}).

We hypothesize the model can minimize the contrastive loss by only modeling the two immovable objects. Since the baseline strategy rarely selects states from the same episode, it suffices to detect whether the current state and the query state come from a different episode. There are 600 unique placements of the two immovable objects, which is enough to distinguish two episodes with high probability.

By using C-SWM-ER with $\beta=0.5$ to include negative states from the same episode as the current state, we can force C-SWM to learn the correct object representations. The 10 step prediction score for 2D Shapes increases to 95\% and for 3D Cubes to 74\% (Table \ref{tab:ap:all_1}, rows 2 and 4). Figure \ref{fig:shapes_latent}b shows the model captures the positions of immovable objects as well as the transitions of movable objects.

\subsection{Exploiting Temporal Correlations in Atari}
\label{sec:exp:atari_orig}

\begin{figure}[h]
    \centering
    \includegraphics[width=1.0\linewidth]{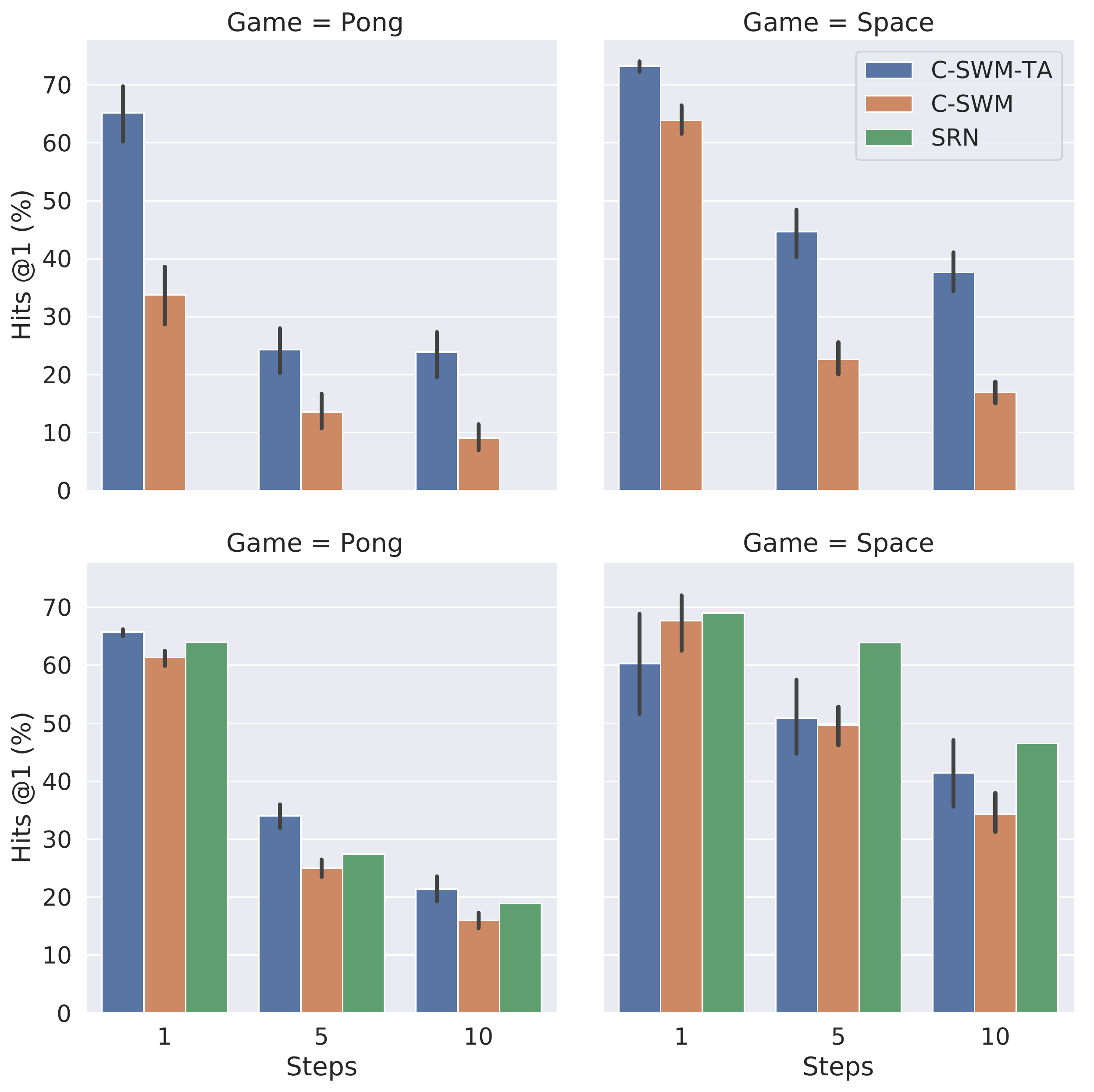}
    \vspace{-2em}
    \caption{First row: comparison between sampling strategies in C-SWM in Pong (first column) and Space Invaders (second column) datasets from \cite{kipf20}. Second row: comparison with the Set Refiner Network (SRN, \cite{huang20better}), which uses different hyper-parameters than \citet{kipf20}. We report means and 95\% confidence intervals over 20 random seeds.}
    \label{fig:cswm_orig_bsr}
\end{figure}

\textbf{Setup.} A small change in negative sampling can also make a large difference in Atari datasets. \citet{kipf20} evaluated C-SWM on Pong and Space Invaders datasets, where an agent takes 10 random actions from the start of the game. Here, states within the same time step of different episodes are correlated. For example, the agent could not have moved the spaceship far away from its starting position at time step 2 in Space Invaders.

\textbf{Result.} Time-aligned negative sampling (C-SWM-TA), which selects from a different episode but at the same time step, doubles the 10 step Hits @1 score for C-SWM from \citet{kipf20} (Figure \ref{fig:cswm_orig_bsr}, 1st row). In subsequent work, \citet{huang20better} found better hyper-parameters for C-SWM in Atari. We report the comparison with updated hyper-parameters in the second row of Figure \ref{fig:cswm_orig_bsr}; we also include the Set Refiner Network \cite{huang20better}. C-SWM-ER performs better in Pong, whereas the Set Refiner Network performs better in Space Invaders. Both differences are statistically significant under the Welch's t-test with $P < 0.001$ (for 10 step prediction). The updated hyper-parameters also decrease the gap between the baseline C-SWM and C-SWM-ER, but differences are statistically significant for Pong with $P < 0.001$, while $P < 0.1$ for Space Invaders (10 step prediction).

\subsection{Training on Diverse Atari Datasets}
\label{sec:exp:atari_full}

\begin{figure}[h]
    \centering
    \includegraphics[width=1.0\linewidth]{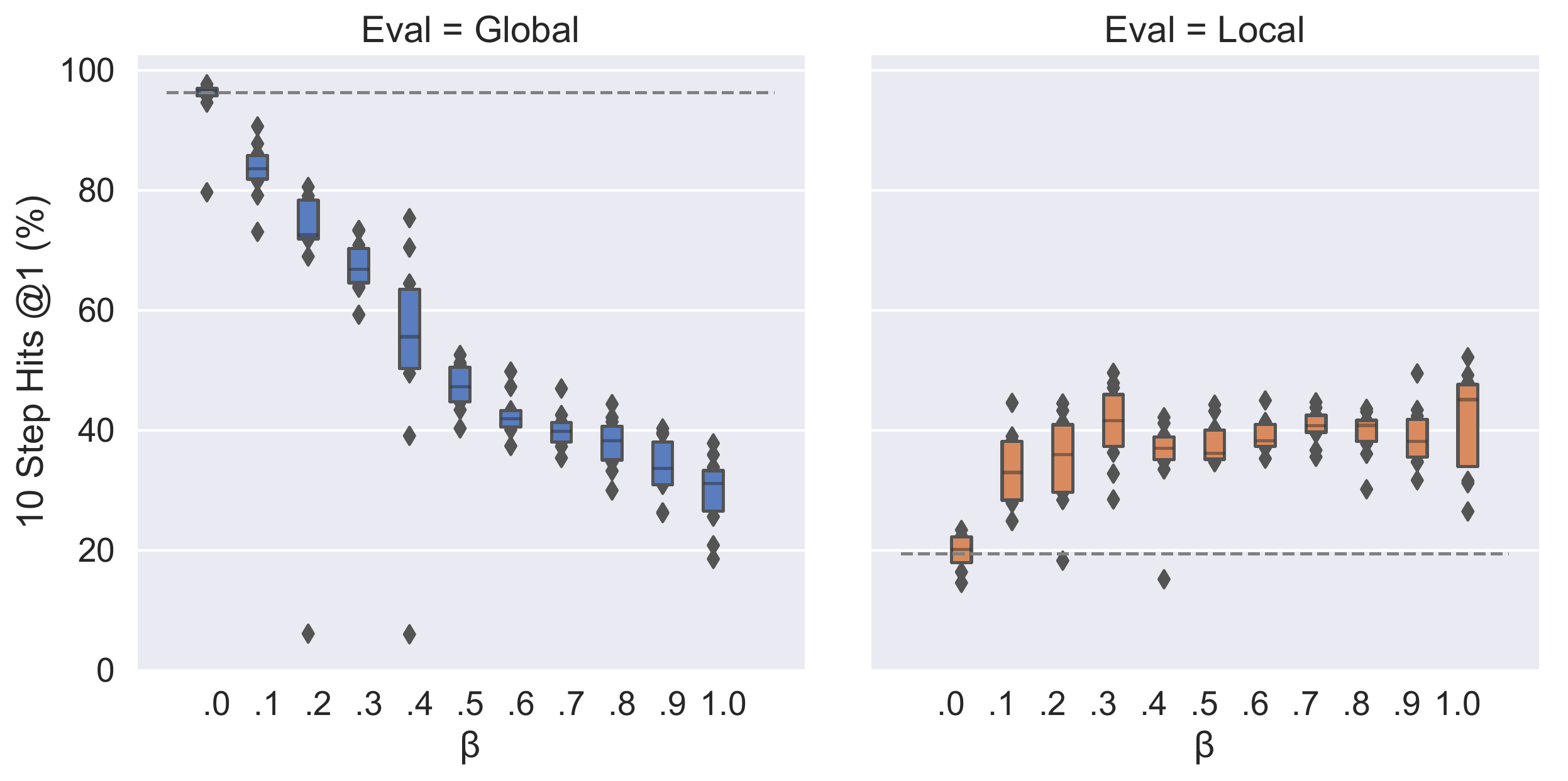}
    \vspace{-2em}
    \caption{Grid search over negative sampling parameter $\beta$ in C-SWM-ER for the Full Pong and Space Invaders datasets. First column: global evaluation metric, second column: local evaluation metrics. Baseline C-SWM performance is denoted by the grey line. We run 10 different random seeds.}
    \label{fig:cswm_full}
\end{figure}

\begin{figure}[h]
    \centering
    \begin{subfigure}{0.32\linewidth}
        \centering
        \includegraphics[width=1.0\linewidth]{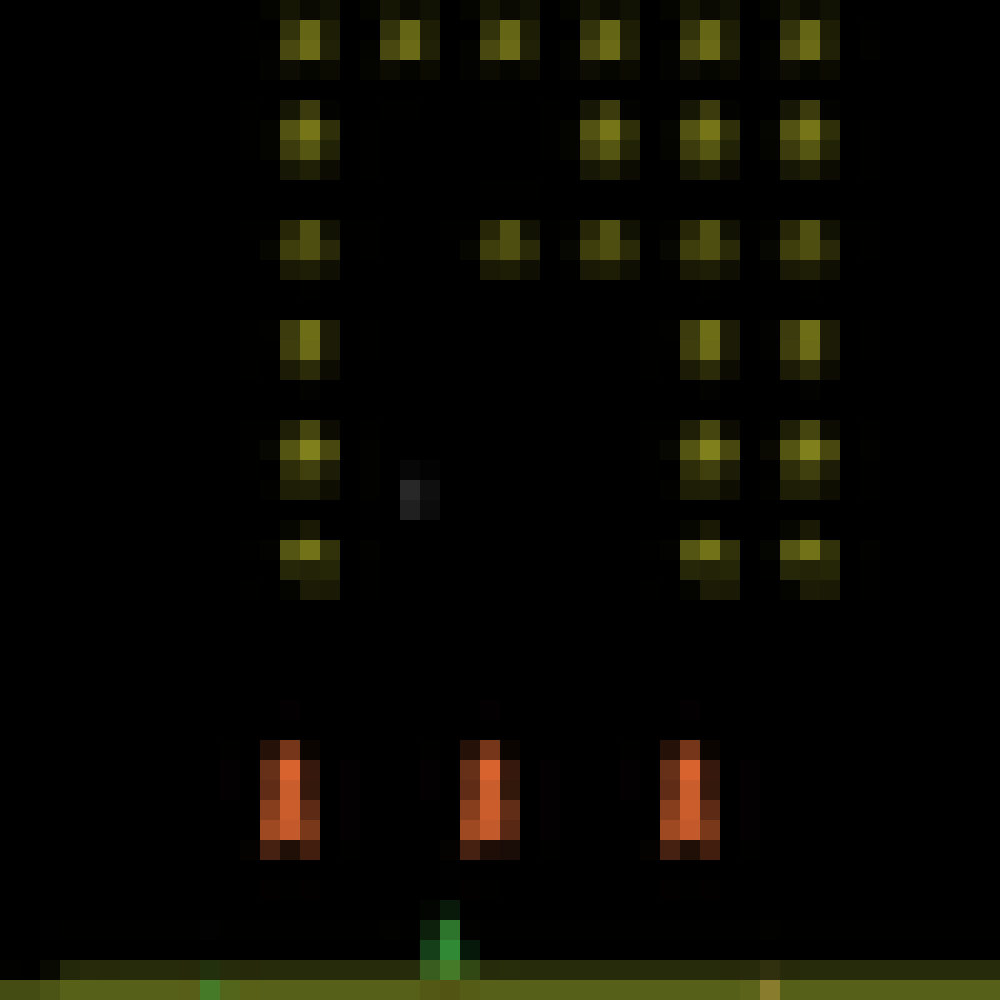}
        \caption{}
    \end{subfigure}
    \begin{subfigure}{0.32\linewidth}
        \centering
        \includegraphics[width=1.0\linewidth]{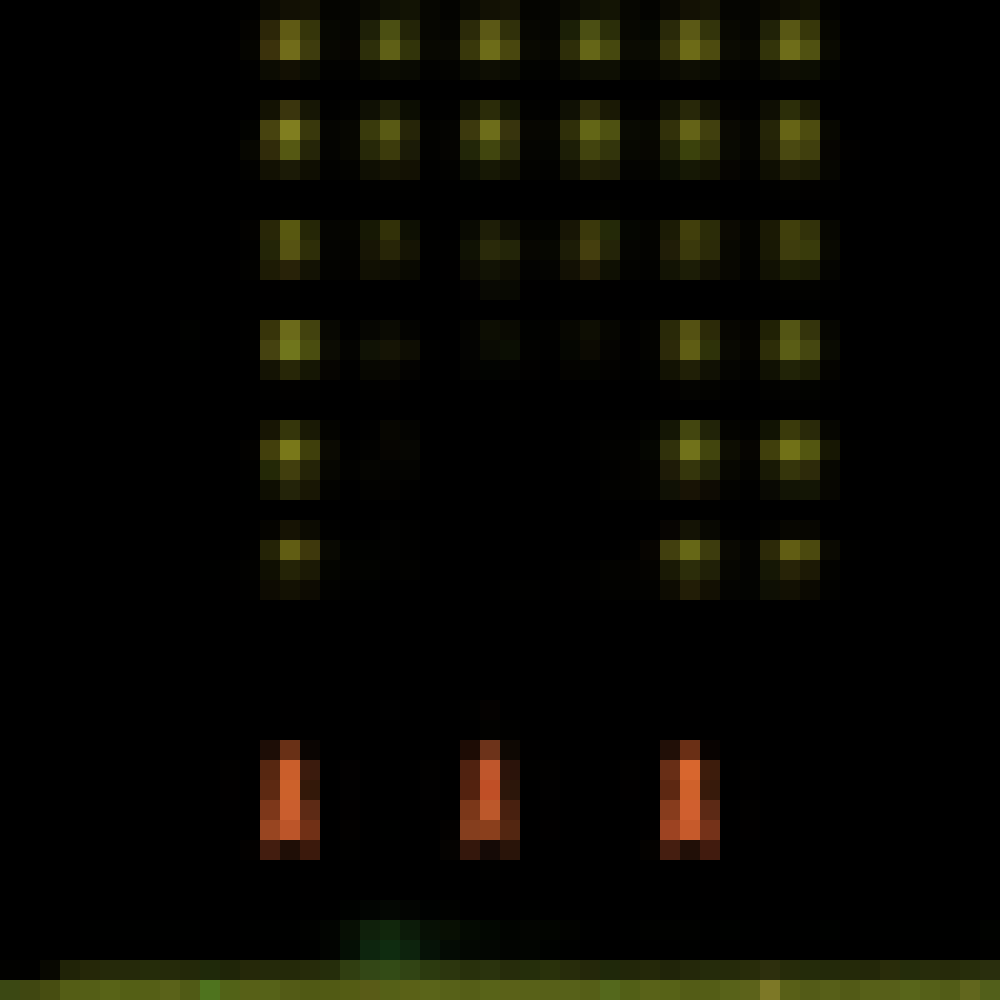}
        \caption{}
    \end{subfigure}
    \begin{subfigure}{0.32\linewidth}
        \centering
        \includegraphics[width=1.0\linewidth]{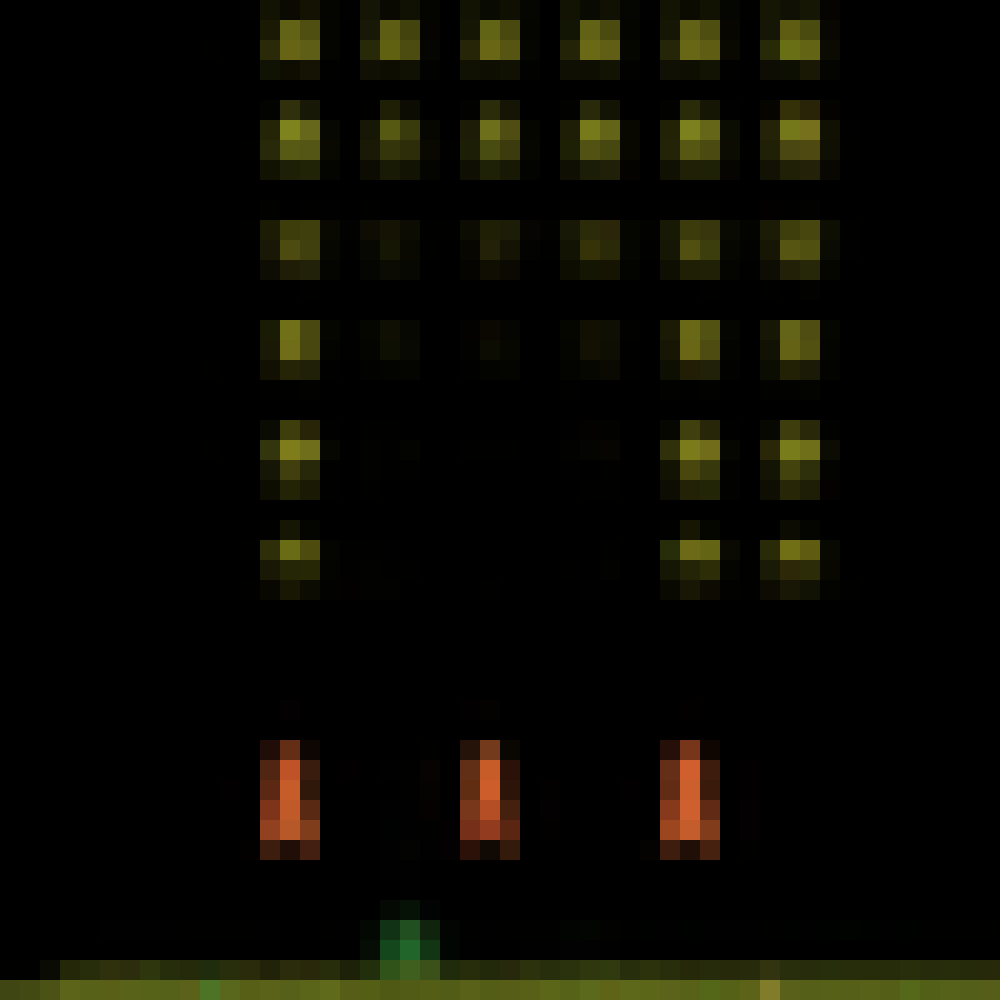}
        \caption{}
    \end{subfigure}
    \begin{subfigure}{0.49\linewidth}
        \centering
        \includegraphics[width=0.49\linewidth]{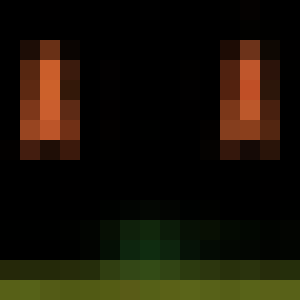}
        \includegraphics[width=0.49\linewidth]{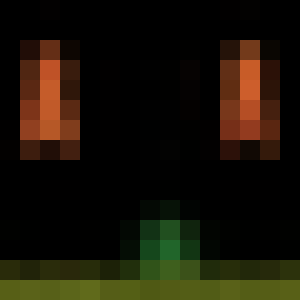}
        \caption{}
    \end{subfigure}
    \begin{subfigure}{0.49\linewidth}
        \centering
        \includegraphics[width=0.49\linewidth]{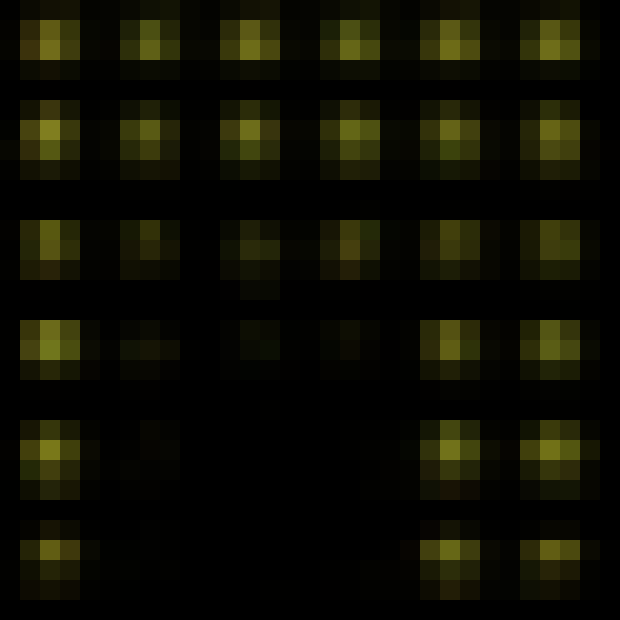}
        \includegraphics[width=0.49\linewidth]{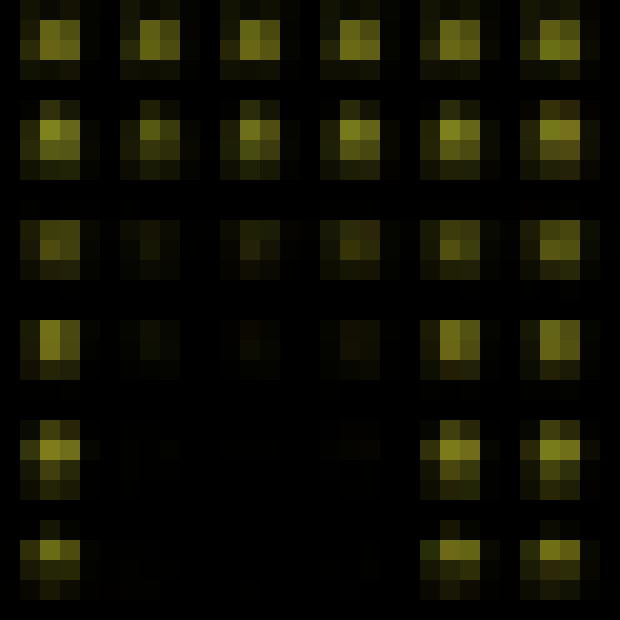}
        \caption{}
    \end{subfigure}
    \caption{Images generated by a decoder trained to reconstruct images based on the latent space of a trained C-SWM with frozen weights. (a) original image at the end of a 10-step episode. (b) and (c) are reconstructed predicted next states (10 steps into the future) using (b) baseline C-SWM and (c) $\beta=0.5$ C-SWM. (d) and (e) are zoomed-in images, where the left one is for the baseline and the right one is for $\beta=0.5$.}
    \label{fig:recons}
\end{figure}

\textbf{Setup.} We break the time-step correlations present in Section \ref{sec:exp:atari_orig} by diversifying the starting states of the Pong and Space Invaders datasets. First, we train A3C, a model-free RL agent, until convergence \cite{mnih16async}. Then, we roll out the trained agent for a random number of steps (see Section \ref{sec:ap:details} for details) with an ($\epsilon=0.5$)-greedy exploration policy. Finally, we take 10 random actions and collect the transitions associated with these 10 steps.

The motivation behind this dataset collection method is to avoid bias in action selection while increasing the diversity of encountered states. Since the dataset does not contain experiences from the trained A3C, we do not have to worry about action bias. We also break the correlation between states encountered at the same time step in different episodes (within the trajectory of 10 random actions), as each episode starts at a different point in the game.

\textbf{Result.} While we would expect the diverse Atari datasets to be more difficult, the 10 step Hits @1 scores are in fact higher than for the original dataset (9\% vs 42\% in Pong and 17\% vs 97\% in Space Invaders; Table \ref{tab:ap:all_3}). 

We hypothesize that the ranking evaluation metric becomes easier as the diversity of the dataset increases. During evaluation, the model has to guess the correct next state (N step into the future) from a batch of 1 correct and 99 incorrect states sampled from different episodes. The more distinct the batch of states is, the easier the task is.

To provide a more balanced view of the model performance, we introduce a modified evaluation metric, which we call Hits @1 Local (L). When predicting the next state $N$ steps into the future, the model has to distinguish the correct next state compared to other states within the same episode.

In Figure \ref{fig:cswm_full}, we compare the baseline C-SWM and C-SWM-ER with varying $\beta$. While increasing $\beta$ decreases the accuracy of the model in the original (Global) evaluation metric, it increases the ability of the model to distinguish within-episode states in Space Invaders. We attempt to evaluate the representation learned by each of the models by trained a decoder on top of a frozen C-SWM latent space (Figure \ref{fig:recons}). The decoder for C-SWM-ER with $\beta=0.5$ seems to be more certain about the position of the spaceship, and decoders for both models seem to have a vague understanding of which aliens have been destroyed. A limitation of this visualization is that we cannot separate decoder and representation failures. Full results are reported in Table \ref{tab:ap:all_3}.

\section{Related Work}
\label{sec:related_work}
\paragraph{Contrastive Representation Learning.} 
Contrastive objectives are widely employed for learning representations of images~\citep{oord2018representation,dosovitskiy2014discriminative,hjelm2018learning,chen2020simple}, natural language~\citep{mnih2012fast,mikolov2013efficient}, graphs~\citep{bordes2013translating,grover2016node2vec,kipf2016variational}, and dynamic environments~\citep{sermanet2018time,kipf20, vdp20, srinivas2020curl}. Apart from the choice of model architecture and design of the loss function, methods primarily differ in how positive and negative samples are obtained. For representation learning on images, positive examples are frequently obtained via data augmentation~\citep{dosovitskiy2014discriminative,hjelm2018learning,chen2020simple}.
For dynamic environments and graphs the (temporal) relationships between the data points provide a natural source for positive pairs in the contrastive learning framework. 

The choice of negative samples can have a substantial impact on performance, and frequently methods choose to sample hard negatives~\citep{oord2018representation,sermanet2018time} that are closely related to the reference sample in order to improve representation quality. In our work, we make a similar observation and find that representations of the recent C-SWM model~\citep{kipf20} can be substantially improved by prioritizing harder negative samples.

\paragraph{Object-centric Models.}
The C-SWM model~\citep{kipf20}, which we study in our work, uses a factorized latent space that represents the state of the environment as a set of latent variables. This approach is inspired by a recent line of work on \textit{object-centric learning} (see \citet{greff2020binding} for a review) which aims to decompose and represent natural input signals, such as images or videos, in terms of separate object components, which facilities dynamics prediction~\citep{watters2017visual}, reasoning~\citep{ding2020object}, and compositional generalization~\citep{van2019perspective}.

\section{Conclusion}

In this paper, we studied three different cases where changes in how we sample negative states in C-SWM have a surprisingly large impact on its performance. We performed experiments on the grid-world and Atari datasets from \citet{kipf20}. We also created a new dataset with better coverage of possible states in Atari Pong and Space Invaders.

%While the transition model of C-SWMs is symmetric under permutation of the latent variables, its encoder produces a fixed ordering, which limits its ability to generalize to objects of novel appearance and to different numbers of objects at test time. Using a symmetry-preserving encoder such as Slot Attention~\citep{locatello2020object,lowe2020learning} for contrastive, object-centric models is an interesting avenue for future work.

\section*{Acknowledgments}

This work was supported by the Intel Corporation, the 3M Corporation, National  Science  Foundation  (1724257, 1724191, 1763878, 1750649, 1835309), NASA (80NSSC19K1474), startup funds from Northeastern University, the Air Force Research Laboratory (AFRL), and DARPA.

\bibliography{main}
\bibliographystyle{icml2021}

\clearpage
\appendix
\section{Additional Data}
\label{sec:ap:additional}

We report results for C-SWM, our negative sampling variants and baseline for varying numbers of object slots (\#Slots). Table \ref{tab:ap:all_1} reports results on 2D Shapes, 3D Cubes and Atari datasets from \citet{kipf20}. Table \ref{tab:ap:all_2} compares our sampling strategy to Set Refiner Networks \cite{huang20better} with their hyper-parameter settings (Section \ref{sec:exp:atari_orig}). Table \ref{tab:ap:all_3} contains results for the full Atari datasets (Section \ref{sec:exp:atari_full}), include an autoencoder-based World Mode baseline.

\begin{figure*}[h]
    \centering
    \begin{subfigure}{1.0\linewidth}
        \centering
        \includegraphics[width=1.0\linewidth]{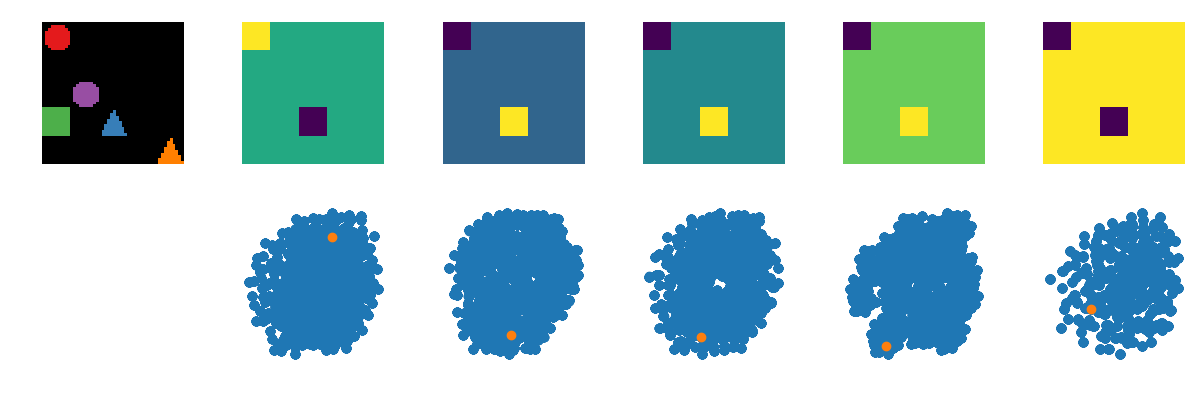}
        \caption{}
    \end{subfigure}
    \begin{subfigure}{1.0\linewidth}
        \centering
        \includegraphics[width=1.0\linewidth]{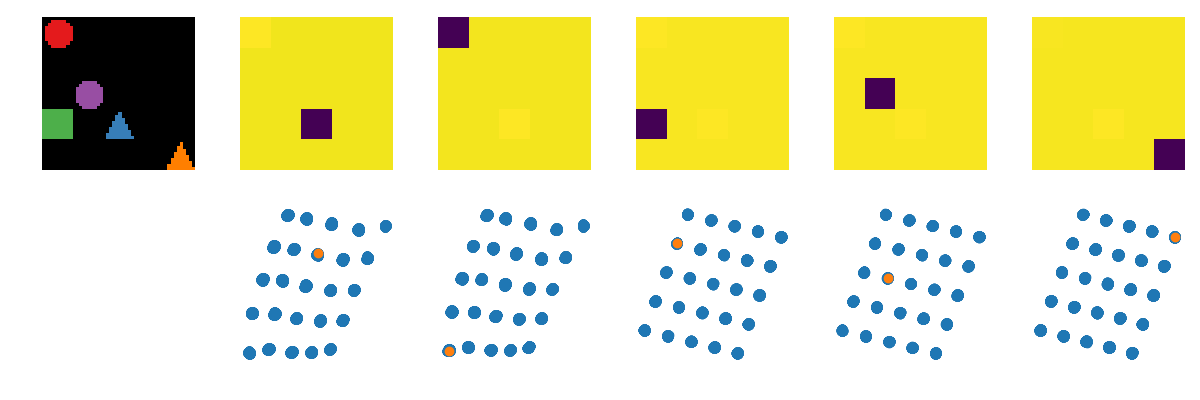}
        \caption{}
    \end{subfigure}
    \caption{Visualization of object extractor activations (first row, second to sixth column) for a random state drawn from the validation set (first column). The activations are $5{\times}5$ feature maps predicted by the object extractor convolutional network. They are scaled between 0 and 1. In the second row, we plot outputs of the object encoder (2D embedding for each object). Blue points are all states in the validation dataset, orange points correspond to the particular state plotted in the first column. The two models are (a) baseline C-SWM and (b) C-SWM-ER with $\beta=0.5$.}
    \label{fig:shapes_latent_slots}
\end{figure*}

We also include an additional visualization of C-SWM-ER ($\beta=0.5$) trained on 2D Shapes, showing how it processes a single state (Figure \ref{fig:shapes_latent_slots}).

\section{Varying Number of Negative Examples}

In this section, we experiment with increasing the number of negative examples, as suggested by our anonymous reviewer. The original C-SWM uses exactly one negative example for each positive example. In our implementation, we sample a batch of negative examples, which is separate from the batch of positive examples. Then, we use \textit{all} samples in the negative batch as negative examples for each positive example. The final loss is an average over the hinge losses of a positive example paired with all negative examples.

Our approach is similar to \citet{vdp20}, except we compare negative examples to the current state, whereas they compare negative examples to the predicted next state. 

\begin{figure}[t!]
    \centering
    \includegraphics[width=\columnwidth]{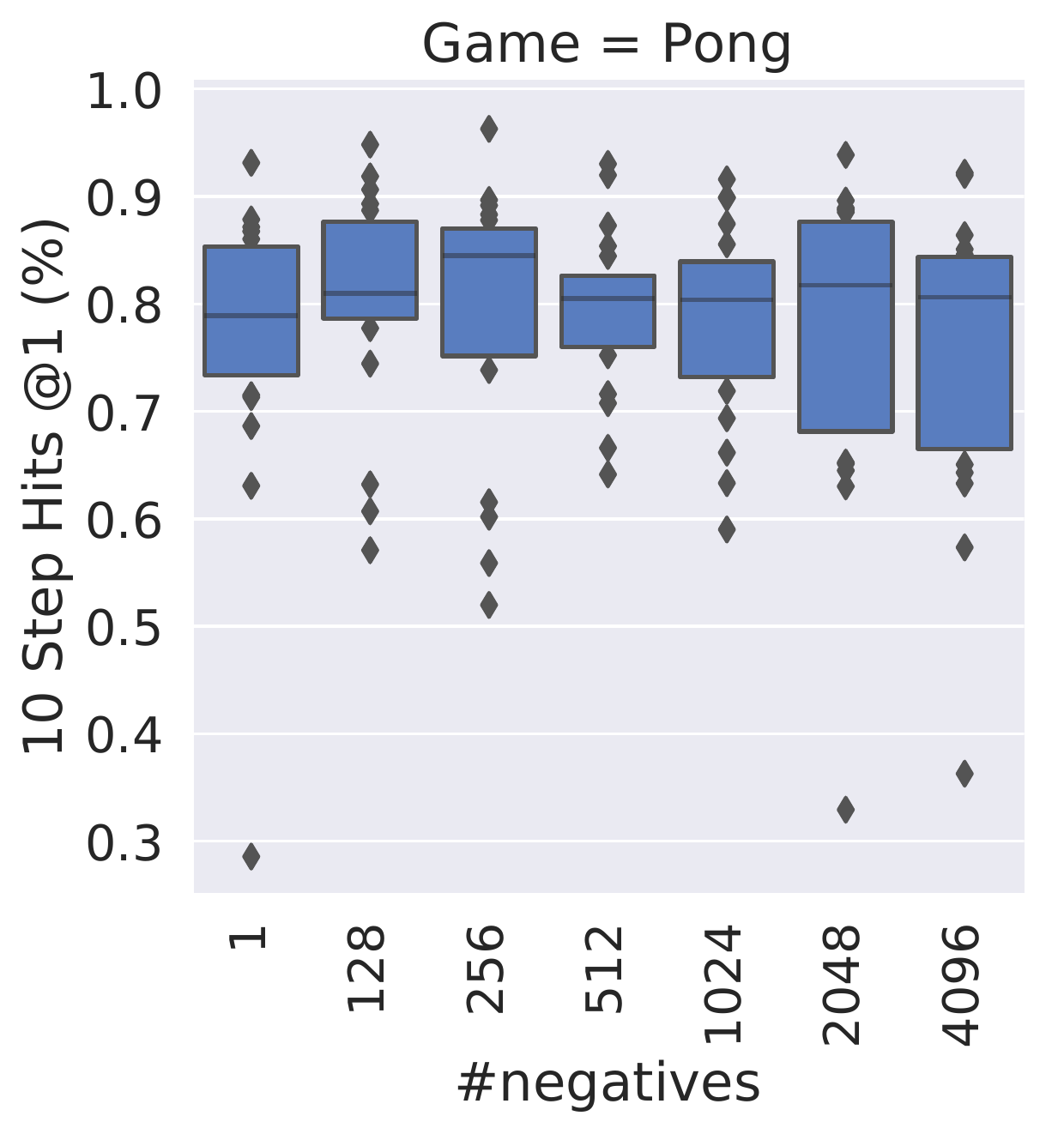}
    \caption{10 step prediction scores of C-SWMs trained with varying number of negative examples on the Full Pong dataset. Each setting was run with 20 random seeds. The model uses three object slots.}
    \label{fig:pong_many_negs}
\end{figure}

Figure \ref{fig:pong_many_negs} shows the results of "global" 10 step evaluation for the Full Pong dataset. There is not a significant difference between 1 and 4096 negative examples. In Space Invaders, the baseline accuracy is already close to perfect (96\%), and our preliminary experiments showed no difference with many negative examples.

Note the baseline C-SWM (1 negative) score is higher than previously reported results (77\% vs 51\%; Figure \ref{fig:cswm_full}, Table \ref{tab:ap:all_3}). We achieved the higher score by decreasing the batch size (to accommodate more negative examples) and running a grid search over learning rates. We used disjoint validation and test sets to prevent overfitting.

\section{Experiment Details}
\label{sec:ap:details}

In this section, we summarize the details of experiments that were based on \citet{kipf20} and describe the details of the full Atari datasets we collected.

\subsection{2D / 3D Grid Immovable}
\label{sec:ap:details:grid}

The 2D Shapes and 3D Cubes environments are based on the C-SWM source code\footnote{URL: https://github.com/tkipf/c-swm, visited on 21/06/17.} with our modifications. We collect 1k episodes of length 100 for training and 10k episodes of length 1k for evaluation. Actions are selected at random. All models are trained for 100 epochs with a batch size of 1024 and a learning rate of $5*10^{-5}$. The neural network weights are initialized using the so-called Xavier initialization with a uniform distribution \cite{glorot10}. The two parameters of the loss function are $\gamma=1.0$ and $\sigma=0.5$. The embedding size of each object slot is 2, resulting in a 10D representation for 5 objects.

\subsection{Original Atari Datasets}
\label{sec:ap:details:atari_orig}

For both Pong and Space Invaders, we collect 1k training episodes and 100 testing episodes with length 10. Actions are selected at random. In the comparison with \citet{huang20better}, we collect a validation set of size 1k (per their evaluation protocol), but we evaluate in batches of 100. That is, for each state, C-SWM needs to distinguish the correct next state embedding ($N$ steps in the future) from 99 other state embeddings. We use an embedding dimension of 4 for each object slot, and the model is trained for 200 epochs. Other hyper-parameters are the same as in Section \ref{sec:ap:details:grid}).

\subsection{Full Atari Datasets}
\label{sec:ap:details:atari_full}

We use an open-source implementation of A3C \cite{mnih16async}\footnote{URL: https://github.com/greydanus/baby-a3c, visited on 21/06/17.} that we train for 40M frames with the default settings. During dataset collection, we use an $\epsilon$-greedy policy with $\epsilon=0.5$. We roll out A3C for a random number of steps sampled between 58 and 100 for Pong and 50 and 300 for Space Invaders. The minimum number of steps is not 0 because the game takes some number of frames to start. After A3C is rolled out for the randomly selected number of steps (which are not included in the dataset), we take 10 random actions and add the corresponding transitions to the dataset. If the game ends during the sequence of 10 random actions we discard the episode. In Space Invaders, the game ends when the spaceship gets shot (we only allow one life). In Pong, we include the experience of the player winning or losing (the ball flying out of the screen), after which both paddles disappear until the end of the episode.

The training dataset contains 10k episodes and the validation dataset 1k. We evaluate in batches of 100 to be comparable to the other Atari experiments. We train with the same settings as in Section \ref{sec:ap:details:atari_orig}; we reduce the number of epochs to 100.

\subsection{Model Architectures}

We use the three types of object extractor networks included in \citet{kipf20}:

\begin{itemize}
    \item Small encoder (2D Shapes): Conv2D ($10{\times}10$ filter, stride 10)-BatchNorm-ReLU-Conv2D($1{\times}1$ filter, stride 1)-Sigmoid.
    \item Medium encoder (Atari): Conv2D ($9{\times}9$ filter, stride 10, padding 4)-BatchNorm-LeakyReLU-Conv2D ($5{\times}5$ filter, stride 5)-Sigmoid.
    \item Large encoder (3D Cubes): Conv2D ($3{\times}10$ filter, padding 1)-BatchNorm-ReLU-Conv2D ($3{\times}10$ filter, padding 1)-BatchNorm-ReLU-Conv2D ($3{\times}10$ filter, padding 1)-Sigmoid.
\end{itemize}

Object encoder uses three FC layers with a hidden layer size of 512 and output size corresponding to embedding size (it is applied separately to each object slot). The FC layers and iterleaved with LayerNorm and ReLU activations. Both the node and edge MLPs in the Graph Neural Networks use the same architecture as object encoder. The edge embedding size is 512.

% table style from the original C-SWM paper
\begin{table*}[ht]
\centering
\caption{\label{tab:ap:all_1}Ranking results for multi-step prediction in latent space. We use model architectures and hyper-parameters from \citet{kipf20}. We report means and standard deviations over 20 random seeds. Our proposed negative sampling strategies are highlighted in bold.}

\begin{adjustbox}{center,width=0.9\linewidth}
\begin{tabular}{lllccccccc}
\toprule
  & & & & \multicolumn{2}{c}{1 Step} & \multicolumn{2}{c}{5 Steps} & \multicolumn{2}{c}{10 Steps} \\ \cmidrule(lr){5-6} \cmidrule(lr){7-8} \cmidrule(lr){9-10}
&&Model & \#Slots & H@1 & MRR & H@1 & MRR & H@1 & MRR      \\ \midrule
\parbox[t]{0.1mm}{\multirow{4}{*}{\rotatebox[origin=c]{90}{2D/3D GRID}}} & \parbox[t]{2mm}{\multirow{4}{*}{\rotatebox[origin=c]{90}{IMMOVABLE}}}&{C-SWM (2D)} & 5 & 53.5{\color{lightgrey}\tiny$\pm$29.6} & 66.3{\color{lightgrey}\tiny$\pm$23.5} & 22.1{\color{lightgrey}\tiny$\pm$19.8} & 37.9{\color{lightgrey}\tiny$\pm$19.3} & 14.0{\color{lightgrey}\tiny$\pm$11.8} & 30.0{\color{lightgrey}\tiny$\pm$12.9}\\
&&{\textbf{C-SWM-ER} ($\beta=0.5$, 2D)} & 5 & {\bf 99.8}{\color{lightgrey}\tiny$\pm$0.3} & {\bf 99.9}{\color{lightgrey}\tiny$\pm$0.2} & {\bf 98.3}{\color{lightgrey}\tiny$\pm$2.3} & {\bf 98.9}{\color{lightgrey}\tiny$\pm$1.6} & {\bf 95.8}{\color{lightgrey}\tiny$\pm$5.9} & {\bf 97.0}{\color{lightgrey}\tiny$\pm$4.4}\\
&&{C-SWM (3D)} & 5 & 61.6{\color{lightgrey}\tiny$\pm$17.2} & 73.3{\color{lightgrey}\tiny$\pm$12.9} & 16.2{\color{lightgrey}\tiny$\pm$10.9} & 32.4{\color{lightgrey}\tiny$\pm$10.4} & 9.9{\color{lightgrey}\tiny$\pm$5.5} & 25.3{\color{lightgrey}\tiny$\pm$5.9}\\
&&{\textbf{C-SWM-ER} ($\beta=0.5$, 3D)} & 5 & {\bf 96.3}{\color{lightgrey}\tiny$\pm$2.0} & {\bf 98.0}{\color{lightgrey}\tiny$\pm$1.1} & {\bf 82.1}{\color{lightgrey}\tiny$\pm$11.6} & {\bf 88.4}{\color{lightgrey}\tiny$\pm$8.4} & {\bf 68.2}{\color{lightgrey}\tiny$\pm$18.4} & {\bf 77.1}{\color{lightgrey}\tiny$\pm$15.3}\\
\midrule
\parbox[t]{0.1mm}{\multirow{6}{*}{\rotatebox[origin=c]{90}{ATARI}}} & \parbox[t]{2mm}{\multirow{6}{*}{\rotatebox[origin=c]{90}{PONG}}}&{C-SWM} & 5 & 24.3{\color{lightgrey}\tiny$\pm$10.1}  & 45.2{\color{lightgrey}\tiny$\pm$10.7} & 8.4{\color{lightgrey}\tiny$\pm$5.0}  & 21.7{\color{lightgrey}\tiny$\pm$7.2} & 6.2{\color{lightgrey}\tiny$\pm$3.5}  & 17.1{\color{lightgrey}\tiny$\pm$5.6}\\
&&{\bf C-SWM-TA} & 5 & {59.1}{\color{lightgrey}\tiny$\pm$14.5}  & {74.2}{\color{lightgrey}\tiny$\pm$12.4} & {21.5}{\color{lightgrey}\tiny$\pm$8.6}  & {36.8}{\color{lightgrey}\tiny$\pm$11.0} & {20.0}{\color{lightgrey}\tiny$\pm$10.0}  & {33.2}{\color{lightgrey}\tiny$\pm$11.7}\\
&&{C-SWM} & 3 & 33.8{\color{lightgrey}\tiny$\pm$11}  & 54.2{\color{lightgrey}\tiny$\pm$10.2} & 13.6{\color{lightgrey}\tiny$\pm$6.6}  & 29.4{\color{lightgrey}\tiny$\pm$8.3} & 9.1{\color{lightgrey}\tiny$\pm$5.0}  & 21.5{\color{lightgrey}\tiny$\pm$6.7}\\
&&{\bf C-SWM-TA} & 3 & {65.2}{\color{lightgrey}\tiny$\pm$10.9}  & {78.8}{\color{lightgrey}\tiny$\pm$7.1} & {24.3}{\color{lightgrey}\tiny$\pm$8.2}  & {42.0}{\color{lightgrey}\tiny$\pm$9.6} & {23.9}{\color{lightgrey}\tiny$\pm$8.7}  & {38.0}{\color{lightgrey}\tiny$\pm$11.0}\\
&&{C-SWM} & 1 & {27.2}{\color{lightgrey}\tiny$\pm$9.7}  & {47.3}{\color{lightgrey}\tiny$\pm$9.8} & {12.4}{\color{lightgrey}\tiny$\pm$4.7}  & {28.3}{\color{lightgrey}\tiny$\pm$6.9} & {8.9}{\color{lightgrey}\tiny$\pm$4.2}  & {21.6}{\color{lightgrey}\tiny$\pm$7.0}\\
&&{\bf C-SWM-TA} & 1 & {\bf 67.9}{\color{lightgrey}\tiny$\pm$5.2}  & {\bf 80.6}{\color{lightgrey}\tiny$\pm$3.3} & {\bf 36.1}{\color{lightgrey}\tiny$\pm$5.5}  & {\bf 54.4}{\color{lightgrey}\tiny$\pm$4.0} & {\bf 28.2}{\color{lightgrey}\tiny$\pm$5.5}  & {\bf 45.9}{\color{lightgrey}\tiny$\pm$5.1}\\
\midrule
\parbox[t]{0.1mm}{\multirow{6}{*}{\rotatebox[origin=c]{90}{SPACE}}} & \parbox[t]{2mm}{\multirow{6}{*}{\rotatebox[origin=c]{90}{INVADERS}}}&{C-SWM} & 5 & {55.1}{\color{lightgrey}\tiny$\pm$8.8}  & {72.1}{\color{lightgrey}\tiny$\pm$7.6} & {22.8}{\color{lightgrey}\tiny$\pm$7.8}  & {41.6}{\color{lightgrey}\tiny$\pm$9.6} & {16.3}{\color{lightgrey}\tiny$\pm$4.9}  & {32.9}{\color{lightgrey}\tiny$\pm$6.4}\\
&&{\bf C-SWM-TA} & 5 & 63.3{\color{lightgrey}\tiny$\pm$5.9}  & 78.5{\color{lightgrey}\tiny$\pm$4.2} & 43.2{\color{lightgrey}\tiny$\pm$12.4} & 61.3{\color{lightgrey}\tiny$\pm$11.3} & 34.2{\color{lightgrey}\tiny$\pm$11.6}  & 51.9{\color{lightgrey}\tiny$\pm$11.8}\\
&&{C-SWM} & 3 & 63.9{\color{lightgrey}\tiny$\pm$5.7} & 78.4{\color{lightgrey}\tiny$\pm$3.7} & 22.7{\color{lightgrey}\tiny$\pm$6.0}  & 43.9{\color{lightgrey}\tiny$\pm$5.9} & 17.0{\color{lightgrey}\tiny$\pm$4.4} & 35.0{\color{lightgrey}\tiny$\pm$5.4}\\
&&{\bf C-SWM-TA} & 3 & {\bf 73.2}{\color{lightgrey}\tiny$\pm$2.1} & {\bf 84.8}{\color{lightgrey}\tiny$\pm$1.3} & {\bf 44.7}{\color{lightgrey}\tiny$\pm$9.2} & {\bf 63.4}{\color{lightgrey}\tiny$\pm$7.3} & {\bf 37.6}{\color{lightgrey}\tiny$\pm$7.8} & {\bf 55.1}{\color{lightgrey}\tiny$\pm$6.7}\\
&&{C-SWM} & 1 & 41.0{\color{lightgrey}\tiny$\pm$21.5} & 57.4{\color{lightgrey}\tiny$\pm$19.7} & 17.6{\color{lightgrey}\tiny$\pm$10.0}  & 35.4{\color{lightgrey}\tiny$\pm$12.6} & 11.9{\color{lightgrey}\tiny$\pm$6.4}  & 26.9{\color{lightgrey}\tiny$\pm$8.7}\\
&&{\bf C-SWM-TA} & 1 & 53.0{\color{lightgrey}\tiny$\pm$23.4}  & 68.1{\color{lightgrey}\tiny$\pm$19.3} & 29.6{\color{lightgrey}\tiny$\pm$11.9}  & 47.9{\color{lightgrey}\tiny$\pm$12.4} & 23.0{\color{lightgrey}\tiny$\pm$9.5}  & 40.0{\color{lightgrey}\tiny$\pm$10.3}\\
\bottomrule
\end{tabular}
\end{adjustbox}
\end{table*}

\begin{table*}[ht]
\centering
\caption{\label{tab:ap:all_2}Ranking results for multi-step prediction in latent space. We use hyper-parameters from \citet{huang20better}. C-SWM-ER results are averaged over 20 random seeds, SRN over 5. We report means and standard deviations. Our proposed negative sampling strategies are highlighted in bold.}
\begin{adjustbox}{center,width=0.9\linewidth}
\begin{tabular}{lllccccccc}
\toprule
  & & & & \multicolumn{2}{c}{1 Step} & \multicolumn{2}{c}{5 Steps} & \multicolumn{2}{c}{10 Steps} \\ \cmidrule(lr){5-6} \cmidrule(lr){7-8} \cmidrule(lr){9-10}
&&Model & \#Slots & H@1 & MRR & H@1 & MRR & H@1 & MRR      \\ \midrule
\parbox[t]{0.1mm}{\multirow{6}{*}{\rotatebox[origin=c]{90}{ATARI}}} & \parbox[t]{2mm}{\multirow{6}{*}{\rotatebox[origin=c]{90}{PONG}}}&{SRN} & 5 & {64.0}{\color{lightgrey}\tiny$\pm$1.2} & {76.0}{\color{lightgrey}\tiny$\pm$0.8} & {27.5}{\color{lightgrey}\tiny$\pm$2.0} & {43.8}{\color{lightgrey}\tiny$\pm$2.0} & {18.9}{\color{lightgrey}\tiny$\pm$3.0}  & {33.4}{\color{lightgrey}\tiny$\pm$3.5}\\
&&{C-SWM} & 5 & 61.3{\color{lightgrey}\tiny$\pm$2.9} & 74.3{\color{lightgrey}\tiny$\pm$2.1} & 25.0{\color{lightgrey}\tiny$\pm$3.6} & 41.7{\color{lightgrey}\tiny$\pm$3.7} & 16.0{\color{lightgrey}\tiny$\pm$3.1} & 29.4{\color{lightgrey}\tiny$\pm$3.8}\\
&&{\bf C-SWM-TA} & 5 & {65.8}{\color{lightgrey}\tiny$\pm$1.3}  & {77.4}{\color{lightgrey}\tiny$\pm$0.8} & {34.1}{\color{lightgrey}\tiny$\pm$4.8} & {50.5}{\color{lightgrey}\tiny$\pm$4.4} & {21.5}{\color{lightgrey}\tiny$\pm$4.9} & {34.9}{\color{lightgrey}\tiny$\pm$5.9}\\
&&{SRN} & 3 & {63.2}{\color{lightgrey}\tiny$\pm$1.4} & {75.4}{\color{lightgrey}\tiny$\pm$0.5} & {28.4}{\color{lightgrey}\tiny$\pm$3.1} & {45.1}{\color{lightgrey}\tiny$\pm$2.3} & {16.1}{\color{lightgrey}\tiny$\pm$2.9}  & {30.9}{\color{lightgrey}\tiny$\pm$1.9}\\
&&{C-SWM} & 3 & 66.0{\color{lightgrey}\tiny$\pm$4.1} & 77.8{\color{lightgrey}\tiny$\pm$2.6} & 29.1{\color{lightgrey}\tiny$\pm$3.5} & 46.1{\color{lightgrey}\tiny$\pm$3.6} & 19.0{\color{lightgrey}\tiny$\pm$3.6} & 33.3{\color{lightgrey}\tiny$\pm$4.0}\\
&&{\bf C-SWM-TA} & 3 & {\bf 70.7}{\color{lightgrey}\tiny$\pm$1.6}  & {\bf 80.8}{\color{lightgrey}\tiny$\pm$1.1} & {\bf 37.2}{\color{lightgrey}\tiny$\pm$4.1} & {\bf 53.6}{\color{lightgrey}\tiny$\pm$3.8} & {\bf 22.7}{\color{lightgrey}\tiny$\pm$3.2} & {\bf 36.4}{\color{lightgrey}\tiny$\pm$3.7}\\
\midrule
\parbox[t]{0.1mm}{\multirow{6}{*}{\rotatebox[origin=c]{90}{SPACE}}} & \parbox[t]{2mm}{\multirow{6}{*}{\rotatebox[origin=c]{90}{INVADERS}}}&{SRN} & 5 & 64.7{\color{lightgrey}\tiny$\pm$1.1} & 79.0{\color{lightgrey}\tiny$\pm$0.7} & 58.0{\color{lightgrey}\tiny$\pm$5.4} & 74.0{\color{lightgrey}\tiny$\pm$4.2} & 36.6{\color{lightgrey}\tiny$\pm$4.5}  & 56.6{\color{lightgrey}\tiny$\pm$4.3}\\
&&{C-SWM} & 5 & 53.0{\color{lightgrey}\tiny$\pm$13.2} & 69.8{\color{lightgrey}\tiny$\pm$10.8} & 38.5{\color{lightgrey}\tiny$\pm$16.0} & 56.4{\color{lightgrey}\tiny$\pm$15.6} & 22.6{\color{lightgrey}\tiny$\pm$11.4} & 39.7{\color{lightgrey}\tiny$\pm$14.2}\\
&&{\bf C-SWM-TA} & 5 & 57.7{\color{lightgrey}\tiny$\pm$13.9} & 73.3{\color{lightgrey}\tiny$\pm$11.5} & 49.9{\color{lightgrey}\tiny$\pm$13.4} & 66.7{\color{lightgrey}\tiny$\pm$11.0} & 36.1{\color{lightgrey}\tiny$\pm$11.8}  & 54.7{\color{lightgrey}\tiny$\pm$11.6}\\
&&{SRN} & 3 & {\bf 69.0}{\color{lightgrey}\tiny$\pm$1.1} & {\bf 81.6}{\color{lightgrey}\tiny$\pm$0.7} & {\bf 63.9}{\color{lightgrey}\tiny$\pm$4.1} & {\bf 78.1}{\color{lightgrey}\tiny$\pm$2.8} & {\bf 46.5}{\color{lightgrey}\tiny$\pm$10.8} & {\bf 65.2}{\color{lightgrey}\tiny$\pm$9.4}\\
&&{C-SWM} & 3 & 67.7{\color{lightgrey}\tiny$\pm$10.8} & 79.8{\color{lightgrey}\tiny$\pm$8.2} & 49.6{\color{lightgrey}\tiny$\pm$7.6} & 67.3{\color{lightgrey}\tiny$\pm$6.1} & 34.3{\color{lightgrey}\tiny$\pm$7.7} & 54.4{\color{lightgrey}\tiny$\pm$7.5}\\
&&{\bf C-SWM-TA} & 3 & 60.3{\color{lightgrey}\tiny$\pm$20.7} & 73.4{\color{lightgrey}\tiny$\pm$16.3} & 50.9{\color{lightgrey}\tiny$\pm$15.2} & 67.5{\color{lightgrey}\tiny$\pm$12.2} & 41.4{\color{lightgrey}\tiny$\pm$13.4}  & 60.3{\color{lightgrey}\tiny$\pm$11.3}\\
\bottomrule
\end{tabular}
\end{adjustbox}
\end{table*}

\begin{table*}[ht]
\centering
\caption{\label{tab:ap:all_3}Ranking results for multi-step prediction in latent space. We report means and standard deviations over 20 random seeds. Our proposed negative sampling strategies are highlighted in bold.}

\begin{adjustbox}{center,width=0.9\linewidth}
\begin{tabular}{lllccccccc}
\toprule
  & & & & \multicolumn{2}{c}{1 Step} & \multicolumn{2}{c}{5 Steps} & \multicolumn{2}{c}{10 Steps} \\ \cmidrule(lr){5-6} \cmidrule(lr){7-8} \cmidrule(lr){9-10}
&&Model & \#Slots & H@1 (G) & H@1 (L) & H@1 (G) & H@1 (L) & H@1 (G) & H@1 (L)      \\ \midrule
\parbox[t]{0.1mm}{\multirow{8}{*}{\rotatebox[origin=c]{90}{FULL ATARI}}} & \parbox[t]{2mm}{\multirow{8}{*}{\rotatebox[origin=c]{90}{PONG}}}&{C-SWM} & 5 & 96.6{\color{lightgrey}\tiny$\pm$4.4} & 98.0{\color{lightgrey}\tiny$\pm$2.1} & {\bf 76.7}{\color{lightgrey}\tiny$\pm$12.3} & 74.5{\color{lightgrey}\tiny$\pm$12.1} & {\bf 52.8}{\color{lightgrey}\tiny$\pm$14.6} & {\bf 67.7}{\color{lightgrey}\tiny$\pm$12.5}\\
&&{\bf C-SWM-ER} ($\beta=0.5$) & 5 & 88.4{\color{lightgrey}\tiny$\pm$3.4} & 98.5{\color{lightgrey}\tiny$\pm$1.0} & 50.7{\color{lightgrey}\tiny$\pm$11.6} & 71.6{\color{lightgrey}\tiny$\pm$11.0} & 27.0{\color{lightgrey}\tiny$\pm$10.5} & 54.5{\color{lightgrey}\tiny$\pm$11.6}\\
&&{C-SWM} & 3 & {\bf 97.5}{\color{lightgrey}\tiny$\pm$1.6} & 96.8{\color{lightgrey}\tiny$\pm$2.3} & 75.0{\color{lightgrey}\tiny$\pm$9.4} & 71.9{\color{lightgrey}\tiny$\pm$8.7} & 50.9{\color{lightgrey}\tiny$\pm$10.8} & 63.0{\color{lightgrey}\tiny$\pm$9.5}\\
&&{\bf C-SWM-ER} ($\beta=0.5$) & 3 & 88.0{\color{lightgrey}\tiny$\pm$2.9} & 98.4{\color{lightgrey}\tiny$\pm$1.1} & 52.9{\color{lightgrey}\tiny$\pm$9.1} & 74.0{\color{lightgrey}\tiny$\pm$8.6} & 26.0{\color{lightgrey}\tiny$\pm$8.5} & 55.1{\color{lightgrey}\tiny$\pm$11.2}\\
&&{C-SWM} & 1 & 96.1{\color{lightgrey}\tiny$\pm$2.1} & 94.0{\color{lightgrey}\tiny$\pm$4.6} & 71.2{\color{lightgrey}\tiny$\pm$7.5} & 70.1{\color{lightgrey}\tiny$\pm$12.5} & 43.3{\color{lightgrey}\tiny$\pm$9.0} & 59.2{\color{lightgrey}\tiny$\pm$10.0}\\
&&{\bf C-SWM-ER} ($\beta=0.5$) & 1 & 95.0{\color{lightgrey}\tiny$\pm$5.3} & {\bf 98.6}{\color{lightgrey}\tiny$\pm$0.8} & 65.7{\color{lightgrey}\tiny$\pm$9.8} & {\bf 74.7}{\color{lightgrey}\tiny$\pm$8.7} & 31.8{\color{lightgrey}\tiny$\pm$8.7} & 54.4{\color{lightgrey}\tiny$\pm$10.9}\\
&&World Model (AE) & 1 & 34.7{\color{lightgrey}\tiny$\pm$19.7} & 56.1{\color{lightgrey}\tiny$\pm$23.5} & 2.9{\color{lightgrey}\tiny$\pm$1.2} & 7.9{\color{lightgrey}\tiny$\pm$1.3} & 1.7{\color{lightgrey}\tiny$\pm$0.7} & 11.1{\color{lightgrey}\tiny$\pm$3.0}\\
\midrule
\parbox[t]{0.1mm}{\multirow{8}{*}{\rotatebox[origin=c]{90}{FULL SPACE}}} & \parbox[t]{2mm}{\multirow{8}{*}{\rotatebox[origin=c]{90}{INVADERS}}}&{C-SWM} & 5 & 99.8{\color{lightgrey}\tiny$\pm$0.1} & 29.4{\color{lightgrey}\tiny$\pm$2.7} & 98.6{\color{lightgrey}\tiny$\pm$0.3} & 17.2{\color{lightgrey}\tiny$\pm$2.4} & {\bf 96.4}{\color{lightgrey}\tiny$\pm$0.8} & 17.5{\color{lightgrey}\tiny$\pm$3.6}\\
&&{\bf C-SWM-ER} ($\beta=0.5$) & 5 & 98.3{\color{lightgrey}\tiny$\pm$0.5} & 77.9{\color{lightgrey}\tiny$\pm$5.1} & 76.3{\color{lightgrey}\tiny$\pm$2.4} & {\bf 44.3}{\color{lightgrey}\tiny$\pm$5.1} & 46.5{\color{lightgrey}\tiny$\pm$4.1} & {\bf 37.7}{\color{lightgrey}\tiny$\pm$4.9}\\
&&{C-SWM} & 3 & {\bf 99.9}{\color{lightgrey}\tiny$\pm$0.1} & 30.5{\color{lightgrey}\tiny$\pm$3.5} & {\bf 98.7}{\color{lightgrey}\tiny$\pm$0.3} & 16.7{\color{lightgrey}\tiny$\pm$1.9,} & 96.3{\color{lightgrey}\tiny$\pm$0.7} & 18.4{\color{lightgrey}\tiny$\pm$3.9,}\\
&&{\bf C-SWM-ER} ($\beta=0.5$) & 3 & 98.9{\color{lightgrey}\tiny$\pm$0.7} & {\bf 80.9}{\color{lightgrey}\tiny$\pm$3.1} & 78.0{\color{lightgrey}\tiny$\pm$4.5} & 41.4{\color{lightgrey}\tiny$\pm$4.4} & 54.7{\color{lightgrey}\tiny$\pm$6.5} & 37.4{\color{lightgrey}\tiny$\pm$4.2}\\
&&{C-SWM} & 1 & {\bf 99.9}{\color{lightgrey}\tiny$\pm$0.1} & 31.6{\color{lightgrey}\tiny$\pm$2.2,} & 98.0{\color{lightgrey}\tiny$\pm$0.4} & 13.6{\color{lightgrey}\tiny$\pm$1.5,} & 94.2{\color{lightgrey}\tiny$\pm$0.9} & 19.8{\color{lightgrey}\tiny$\pm$4.0,}\\
&&{\bf C-SWM-ER} ($\beta=0.5$) & 1 & 97.2{\color{lightgrey}\tiny$\pm$0.8} & 75.4{\color{lightgrey}\tiny$\pm$5.5} & 74.8{\color{lightgrey}\tiny$\pm$7.4} & 35.3{\color{lightgrey}\tiny$\pm$2.4} & 55.0{\color{lightgrey}\tiny$\pm$9.9} & 36.8{\color{lightgrey}\tiny$\pm$3.0}\\
&&World Model (AE) & 1 & 64.2{\color{lightgrey}\tiny$\pm$20.8} & 12.9{\color{lightgrey}\tiny$\pm$3.1} & 35.8{\color{lightgrey}\tiny$\pm$12.5} & 5.6{\color{lightgrey}\tiny$\pm$1.1} & 28.6{\color{lightgrey}\tiny$\pm$10.3} & 11.8{\color{lightgrey}\tiny$\pm$4.3}\\
\bottomrule
\end{tabular}
\end{adjustbox}
\end{table*}

%%%%%%%%%%%%%%%%%%%%%%%%%%%%%%%%%%%%%%%%%%%%%%%%%%%%%%%%%%%%%%%%%%%%%%%%%%%%%%%
%%%%%%%%%%%%%%%%%%%%%%%%%%%%%%%%%%%%%%%%%%%%%%%%%%%%%%%%%%%%%%%%%%%%%%%%%%%%%%%

\end{document}